\documentclass[preprints,article,accept,oneauthor,pdftex]{Definitions/mdpi}

\firstpage{1} 
\pubvolume{xx}
\issuenum{1}
\articlenumber{5}
\pubyear{2020}
\copyrightyear{2020}
\history{}
\Title{A Generalized Flow for B2B Sales Predictive Modeling: An Azure Machine Learning Approach}

\Author{Alireza Rezazadeh\orcidA{}}

\AuthorNames{Alireza Rezazadeh}

\address[1]{%
Electrical and Computer Engineering Department, University of Illinois at Chicago, Chicago, Illinois, USA; arezaz2@uic.edu\\}

\abstract{Predicting the outcome of sales opportunities is a core part of successful business management. Conventionally, making this prediction has relied mostly on subjective human evaluations in the process of sales decision making. In this paper, we addressed the problem of forecasting the outcome of business to business (B2B) sales by proposing a thorough data-driven Machine Learning (ML) workflow on a cloud-based computing platform: Microsoft Azure Machine Learning Service (Azure ML). This workflow consists of two pipelines: (1) An ML pipeline to train probabilistic predictive models on the historical sales opportunities data. In this pipeline, data is enriched with an extensive feature enhancement step and then used to train an ensemble of ML classification models in parallel. (2) A prediction pipeline to utilize the trained ML model and infer the likelihood of winning new sales opportunities along with calculating optimal decision boundaries. The effectiveness of the proposed workflow was evaluated on a real sales dataset of a major global B2B consulting firm. Our results implied that decision-making based on the ML predictions is more accurate and brings a higher monetary value.}


\begin{document}
%%%%%%%%%%%%%%%%%%%%%%%%%%%%%%%%%%%%%%%%%%

%%%%%%%%%%%%%%%%%%%%%%%%%%%%%%%%%%%%%%%%%%
\section{Introduction}
In the Business to Business (B2B) commerce, companies compete to win high-valued sales opportunities to maximize their profitability.  In this regard, a key factor for maintaining a successful B2B enterprise is the task of forecasting the outcome of sales opportunities. B2B sales process typically demands significant costs and resources and, hence, requires careful evaluations in the very early steps. Quantifying the likelihood of winning new sales opportunities is an important basis for appropriate resource allocation to avoid wasting resources and sustain the company's financial objectives \cite{monat2011industrial, bohanec2016integration, matthies2018double, duran2008probabilistic}.

Conventionally, forecasting the outcome of sales opportunities is carried out mostly relying on subjective human rating \cite{bohanec2017organizational, ingram2015sales, davis2007organizational, armstrong2015golden}. Most of the Customer Relationship Management (CRM) systems allow salespersons to manually assign a probability of winning for new sales opportunities \cite{xu2017hitting}. This probability is then used at various stages of the sales pipeline, i.e., calculating a weighted revenue of the sales records \cite{davis2001determining, de2015decision}. Often each salesperson develops a non-systematic intuition to forecast the likelihood of winning a sales opportunity with little to no quantitative rationale, neglecting the complexity of the business dynamics \cite{xu2017hitting}. Besides, as often as not, sales opportunities are intentionally underrated to avoid any internal competition with other salespersons or overrated to circumvent the pressure from sales management to maintain a higher performance \cite{yan2015sales}.

Even though the abundance of data and improvements in statistical and machine learning (ML) techniques have led to significant enhancements in data-driven decision-making, the literature is scarce in the subject of B2B sales outcome forecasting. Yan et al. \cite{yan2015sales} explored predicting win-propensity of sales opportunities using a two-dimensional Hawkes dynamic clustering technique. Their approach allowed for live assessment of active sales although relied heavily on regular updates and inputs from salespersons in the CRM system. This solution is hard to maintain in larger B2B firms considering each salesperson often handles multiple opportunities in parallel and would put less effort into making frequent interaction with each sales record \cite{lambert2018sales}.

Tang et al. \cite{xu2017hitting} built a sales forecast engine consist of multiple models trained on snapshots of historical data. Although their paradigm is focused on revenue forecasting, they demonstrated the effectiveness of hybrid models for sales predictive modeling. Bohane et al. \cite{bohanec2017organizational} explored the idea of single and double-loop learning in B2B forecasting using ML models coupled with general explanation methods. Their main goal was actively involving users in the process of model development and testing. Built on their earlier work on effective feature selection \cite{bohanec2015feature} they concluded random forest models were the most promising for B2B sales forecasting.

Here, we proposed an end-to-end cloud-based workflow to forecast the outcome of B2B sales opportunities by reframing this problem into a binary classification framework. First, an ML pipeline extracts sales data and improves them through a comprehensive feature enhancement step. The ML pipeline optimally parameterizes a hybrid of probabilistic ML classification models trained on the enhanced sales data and eventually outputs a voting ensemble classifier. Second, a prediction pipeline makes use of the optimal ML model to forecast the likelihood of winning new sales opportunities. Importantly, the prediction pipeline also performs thorough statistical analysis on the historical sales data and specifies appropriate decision boundaries based on sales monetary value and industry segment. This helps to maximize the reliability of predictions by binding the interpretation of model results to the actual data.

The proposed workflow was implemented and deployed to a global B2B consulting firm's sales pipeline using Microsoft Azure Machine Learning Service (Azure ML). Such a cloud-based solution readily integrates into the existing CRM systems within each enterprise and allows for more scalability. Finally, we compared the performance of the proposed solution with salespersons' predictions using standard statistical metrics (e.g. accuracy, AUC, etc.). To make the comparison more concrete, we also looked into the financial aspect of implementing this solution and compared the monetary value of our ML solution with salespersons' predictions. Overall, we have found that the proposed ML solution results in a superior prediction both in terms of statistical and financial evaluations; therefore, it would be a constructive complement to the predictions made by salespersons.

%%%%%%%%%%%%%%%%%%%%%%%%%%%%%%%%%%%%%%%%%%
\section{Materials and Methods}

\subsection{Data and Features} \label{data and feature}
\unskip
\subsubsection{Data}
\begin{figure}[H]
\centering
\includegraphics[width=4 in]{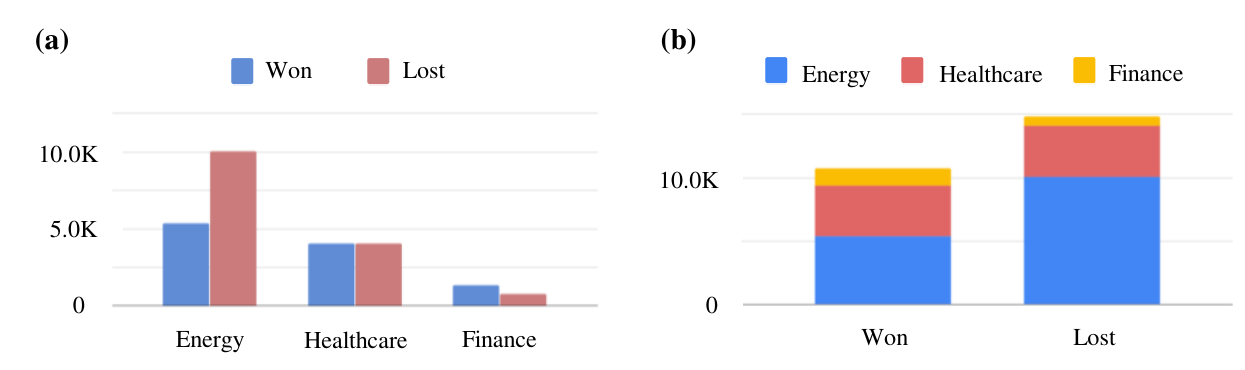}
\caption{Data Exploration: (\textbf{a}) Distribution of sales opportunity records across the three business segments: Healthcare, Energy, and Finance. (\textbf{b}) Closed sales opportunities final status.}
\label{fig: data exploration}
\end{figure}

Data for this study were obtained from a global multi-business B2B consulting firm's CRM database in three main business segments: Healthcare, Energy, and Financial Services (Finance for short). A total number of 25578 closed sales opportunity records starting January 2015 through August 2019 were used in this work (Figure \ref{fig: data exploration}a). Each closed opportunity record contained a status label (won/lost) corresponding to its ultimate outcome, otherwise if still active in the sales pipeline, they were labeled as open. Out of all closed sales records $\sim$58\% were labeled as "won" in their final sales status (Figure \ref{fig: data exploration}b).  

A total number of 20 relevant variables (features) were extracted for each sales opportunity from the raw CRM database. Table \ref{table: features} describes these features in more details. Specifically, a subset of the features described the sales project (Opportunity Type, General Nature of Work, Detailed Nature of Work, Project Location, Project Duration, and Total Contract Value, Status). The remaining features provided further information on the customer (Account, Account Location, Key Account Energy, Key Account Finance, and Key Account Healthcare) and the internal project segmentation and resource allocation (Business Unit, Engagement Manager, Sales Lead, Probability, Sub-practice, Practice, Group Practice, Segment, and User-entered Probability).

\begin{table}
\centering
\caption{Raw CRM sales database features.}
\label{table: features}
\begin{tabular}{lll} 
\toprule
\multicolumn{1}{c}{ \textbf{count} } & \multicolumn{1}{c}{\textbf{Feature Type} } & \multicolumn{1}{c}{\textbf{Features} }                                                                                                                                                                                                                      \\ 
\hline
13                                   & Categorical                                & \begin{tabular}[c]{@{}l@{}}Business Unit, Opportunity Type, Project Location, General Nature of Work,\\ Detailed Nature of Work, Account, Account Location, Sales Lead\\ Engagement Manager, Sub-practice, Practice, Group Practice, Segment \end{tabular}  \\ 
\cmidrule(lr){1-3}
4                                    & Binary                                     & Status, Key Account Energy, Key Account Healthcare, Key Account Finance                                                                                                                                                                                     \\ 
\cmidrule(r){1-3}
3                                    & Continuous                                 & User-entered Probability, Project Duration, Total Contract Value                                                                                                                                                                                            \\
\bottomrule
\end{tabular}
\end{table}

Once a sales opportunity profile was created in the CRM system, users were required to input their estimation for the probability of winning that opportunity. Note that the user-entered probabilities were not used in the process of training ML models and were only used as a point of reference to compare with the performance of the ML workflow. All the features listed in Table \ref{table: features} were required to populate in the CRM system; therefore, less than 1\% of the dataset contained missing values. As a result, sales records with a missing value were dropped from the dataset.

\subsection{Feature Enhancement}\label{Feature Enhancement}
\unskip

The CRM raw dataset was enhanced by inferring additional relevant features calculated across the sales records. These additional features were calculated using statistical analysis on the categorical features: Sales Leads, Account, Account Location, etc. Mainly, the idea was to extract a lookup table containing relevant statistics calculated across the sales records for each of the unique values in the categorical features. 

By collecting the historical data of unique values of each categorical features (i.e, for each individual Sales Lead, Account, and Project Location, etc.), we calculated the following statistical metrics: (1) Total number of sales opportunities (2) Total number of won sales (3) Total number of lost sales (4) Average contract value (value for short) of won sales (5) Standard error of the mean won sales value (6) Winning rate calculated as the ratio of won and total sales counts (7) Coefficient of variation \cite{abdi2010coefficient} of won sales value to capture the extent of variability in the won sales contract values.

\begin{figure}[H]
\centering
\includegraphics[width=5 in]{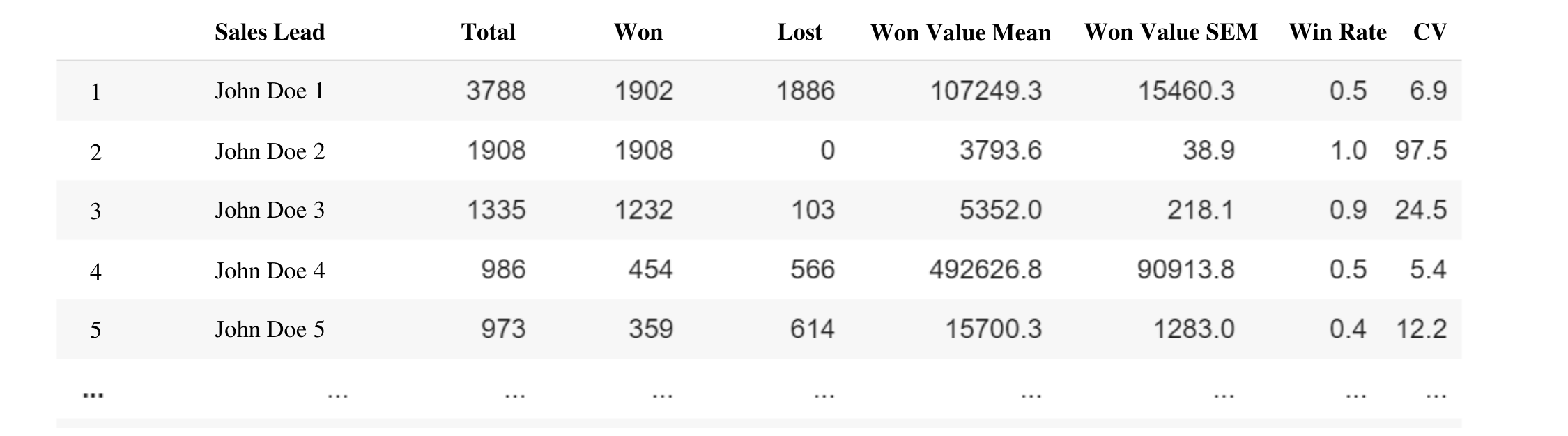}
\caption{Feature Enhancement Lookup Table: An example of the statistics calculated based on “Sales Lead” including counts of the total, won, and lost opportunities along with the mean and standard error of the mean (SEM) for won sales value and their coefficient of variation (CV).}
\label{fig: lookup table}
\end{figure}

The aforementioned statistics were calculated and stored in feature enhancement lookup tables for each categorical features (see Table \ref{table: features} for a list of these features). Figure \ref{fig: lookup table} provides an example of a feature enhancement lookup table calculated based on the "Sales Lead" feature in the raw CRM dataset. These lookup tables (13 tables in total for all categorical features) were appropriately merged back to the raw CRM sales dataset.

In the last feature enhancement step, the Mahalanobis \cite{de2000mahalanobis} distance was calculated between each sales opportunity's value and the distribution of all won sales value that shared a similar categorical feature (individually for each of the 13 categorical features). This quantifies how far a sales value is relative to the family of won sales with the same characteristics (i.e, same Sales Lead, Project Location, Segment, etc.). The process of feature enhancement increased the total number of features to 137 for each sales record (20 features originally from the raw CRM dataset + \(9\times 13 = 117\) additional features from the lookup tables).

The enhanced CRM dataset (25578 total number of sales opportunities) was randomly split into a "train set" (70\%) and a "test set" (30\%). The train set was used to train ML models with a 10-fold cross-validation technique.  The test set was used to report the performance of the trained ML model on the unseen portion of the dataset. For further evaluations, after the proposed framework was deployed to the sales pipeline of the enterprise, a "validation set" was collected of new sales records over a period of 3 months (846 closed sales opportunities).

\subsection{Machine Learning Overview}
\unskip
Our solution to predicting the likelihood of winning sales opportunities is essentially reframing this problem in a supervised binary classification paradigm (won, lost). Hence, we made use of two of the most promising supervised classification algorithms: XGBoost, and LightGBM. Particularly, these two models were selected among the family of probabilistic classification models due to their higher classification accuracy in our problem. A second motivation for using these two models was the fact that the distributed versions of both can easily integrate into cloud platforms such as Azure ML. Last, to attain a superior performance, multiple iterations of both models were combined in a voting ensemble. 

\subsubsection{Binary Classification}\label{binary classification}
Probabilistic classification algorithms \cite{abbott_2018}, given pairs of samples and their corresponding class labels \((X_1, y_1), \dotsc, (X_n, y_n)\), capture a conditional probability distribution over the output classes \(P(y_i\in{Y}\mid X_i)\) where for a binary classification scenario \(Y\in{\{0, 1}\}\) (maps to lost/won in our problem). Given the predicted probability of a data sample, a decision boundary is required to define a reference point and predict which class the sample belongs to. In a standard binary classification, the predicted class is the one that has the highest probability \cite{bishop2006pattern}. This translates to a standard decision boundary of \(0.5\) for predicting class labels.

\begin{figure}[H]
\centering
\includegraphics[width=2.6 in]{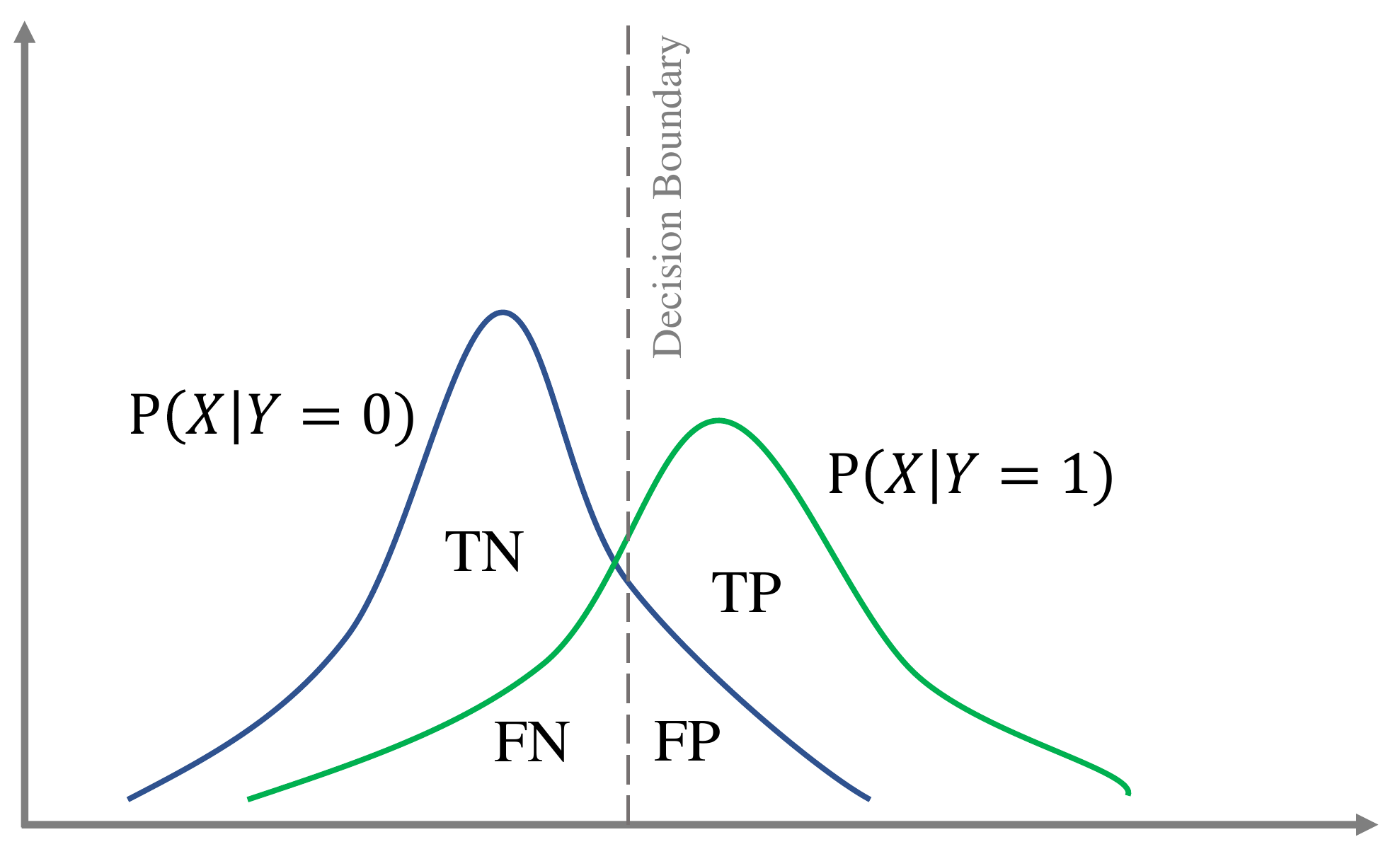}
\caption{Decision Boundary: impact of the decision boundary on different scenarios of the binary classification.}
\label{fig: decision boundary}
\end{figure}

However, the decision boundary can be calibrated arbitrarily to reflect more on the distribution of the data. The influence of the decision boundary on the number of true positives \((TP)\), false positives \((FP)\), true negatives \((TN)\), and false negatives \((FN)\) in binary classification is illustrated in Figure \ref{fig: decision boundary} (see Table \ref{table: metrics} for definitions). In this work, we find the optimal decision boundary for a classification model by maximizing all true conditions (both \(TP\) and \(TN\)) which in return, minimizes all the false conditions (\(FP\) and \(FN\)). Visually, this decision boundary is a vertical line passing through the intersections of \(P(X|Y=0)\) and \(P(X|Y=1)\) in Figure \ref{fig: decision boundary}.

The performance of a binary classifier can be evaluated using standard statistical metrics such as accuracy, precision, recall, and F1-score (see Table \ref{table: metrics}). For the case of binary classification, the area under ROC curve (AUC) measures the robustness of the classification (a higher AUC suggests more robust classification performance) \cite{bewick2004statistics}.

We took a step forward to obtain more insight into the classification results and measured the performance of the classifier from a monetary aspect, that is, we calculated the value created by adopting a classification algorithm in the decision-making process. In particular, we aggregated the total sales values in each of the four scenarios of classification (\(TP_m, TN_m, FP_m, FN_m\)) and defined monetary performance metrics with a similar formulation to the statistical metrics (see Table \ref{table: metrics}). For instance, the monetary precision is the fraction of the sales values correctly predicted as won.

\begin{table}
\centering
\caption{Statistical and monetary classifier performance metrics.}
\label{table: metrics}
    \begin{tabular}{ll}
    \toprule
    \multicolumn{2}{l}{\textit{Statistical Metrics}}                                     \\ 
    \midrule
    \textbf{Notation}     & \textbf{Definition}                                                   \\ 
    \midrule
    \(TP\)           & True
      Positive: number of class 1 samples classified as 1   \\
    \(TN\)           & True Negative: number of class
    0 samples classified as 0     \\
    \(FP\)           & False Positive:
      number of class 0 samples classified as 1  \\
    \(FN\)           & False Negative: number of class
    1 samples classified as 0    \\ 
    \toprule
    \textbf{Metric}       & \textbf{Definition}                                                   \\ 
    \midrule
    Precision    & \(TP/(TP+FP)\)                                                   \\
    Recall       & \(TP/(TP+FN)\)                                                   \\
    Accuracy     & \((TP+TN)/(TP+TN+FP+FN)\)                                        \\
    F1-Score     & \(2/(Recall^{-1} + Precision^{-1})\)                           \\ 
    \toprule
    \toprule
    \multicolumn{2}{l}{\textit{Monetary Metrics}}                                        \\ 
    \midrule
    \textbf{Metric}       & \textbf{Definition}                                                   \\ 
    \midrule
    Precision\textsubscript{m}   & \(TP_m/(TP_m+FP_m)\)                                             \\
    Recall\textsubscript{m}      & \(TP_m/(TP_m+FN_m)\)                                             \\
    Accuracy\textsubscript{m}    & \((TP_m+TN_m)/(TP_m+TN_m+FP_m+FN_m)\)                            \\
    \toprule
    \end{tabular}
\end{table}

\subsubsection{XGBoost and LightGBM Classifiers} \label{XGBoost and LightGBM}

XGBoost, introduced by Chen and Guestrin \cite{chen2016xgboost}, is a supervised classification algorithm that iteratively combines weak base learners into a stronger learner. With this algorithm, the objective function \(J\) is defined as

\begin{equation} \label{eqn1}
    J(\Theta) = L(y, \hat{y}) + \Omega(\Theta)
\end{equation}

where \(\Theta\) denotes the model's hyperparameters. The training loss function \(L\) quantifies the difference between the prediction \(\hat{y}\) and actual target value \(y\). The regularization term \(\Omega\) penalizes the complexity of the model with the L1 norm to smooth the learned model and avoid over-fitting.
The model's prediction is an ensemble of \(k\) decision trees from a space of trees \(\mathcal{F}\):

\begin{equation} \label{eqn2}
    \hat{y}_i = \sum_{i=1}^{k} f_k(x_i) , f_k \in{\mathcal{F}}
\end{equation}

The objective function at iteration \(t\) for \(n\) instances can be simplified as:

\begin{equation} \label{eqn3}
    J^{(t)} = \sum_{i=1}^{n} L(y_i, \hat{y}_i) + \sum_{k=1}^{t} \Omega (f_k)
\end{equation}

where according to Eq (\ref{eqn2}), \(\hat{y}_i\) can iteratively be written as

\begin{equation} \label{eqn4}
    \hat{y_i}^{(t)} = \sum_{i=1}^{t} f_k(x_i) = \hat{y_i}^{(t-1)} + f_t(x_i)
\end{equation}

The regularization term can be defined as

\begin{equation} \label{eqn5}
     \Omega (f_k) = \gamma T + \frac{1}{2}\lambda\sum_{j=1}^{T} w_j^2
\end{equation}

where the coefficient \(\gamma\) is the complexity of each leaf. Also, \(T\) is the total number of leaves in the decision tree. To scale the weight penalization, \(\lambda\) can be tweaked. Using second-order Taylor expansion and assuming a mean square error (MSE) loss function, Eq (\ref{eqn3}) can be written as

\begin{equation} \label{eqn6}
    J^{(t)} \approx \sum_{i=1}^{n} [g_i w_{q(x_i)} + \frac{1}{2}(h_i w_{q(x_i)})^2] +  \frac{1}{2}\lambda\sum_{j=1}^{T} w_j^2
\end{equation}

Since each incident of data corresponds to only one leaf, according to \cite{zhang2018data}, this can also be simplified as

\begin{equation} \label{eqn7}
    J^{(t)} \approx \sum_{j=1}^{T} [(\sum_{i\in{I_j}} g_i) w_j + \frac{1}{2}(\sum_{i\in{I_j}} h_i + \lambda) w_j^2] + \gamma T
\end{equation}

where \(I_j\) represents all instances of data in leaf \(j\). As can be seen in Eq (\ref{eqn7}) minimizing the objective function can be transformed into finding the minimum of a quadratic function. 

\begin{figure}[H]
\centering
\includegraphics[width=3.5in]{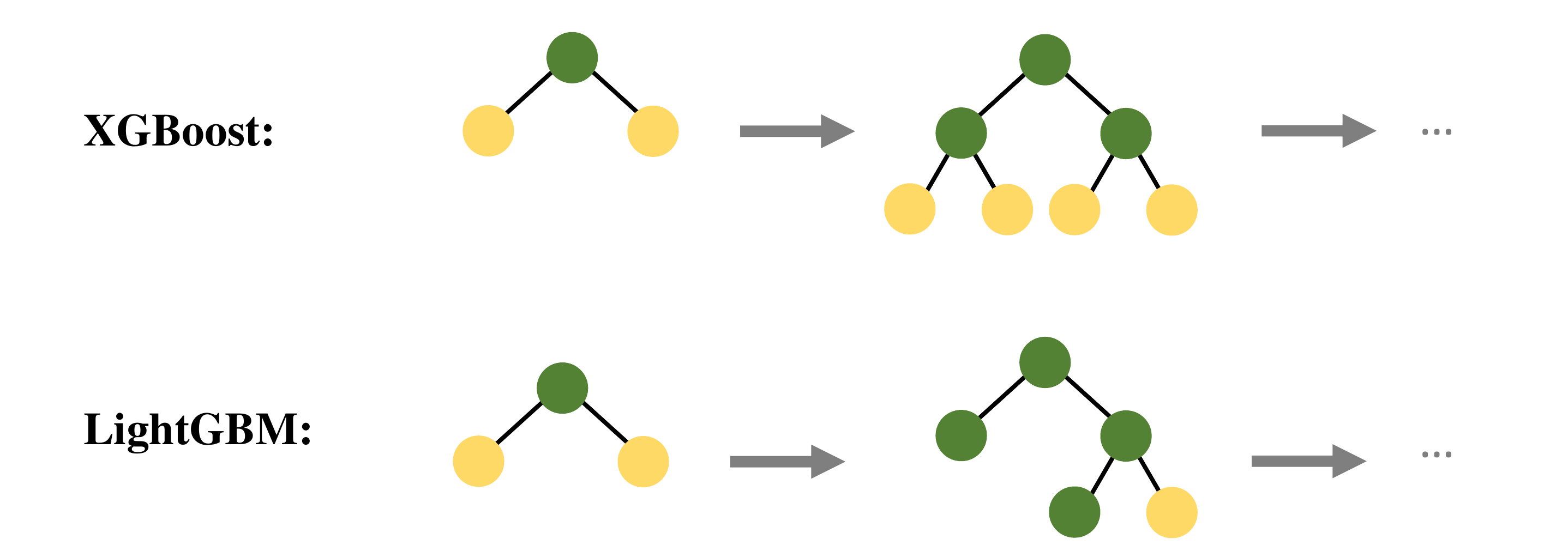}
\caption{Comparison between XGBoost level-wise horizontal tree growth and LightGBM vertical leaf-wist tree growth.}
\label{fig: modelscomparison}
\end{figure}

In an endeavor to reduce the computation time of the XGBoost algorithm, Ke, et al. proposed LightGBM \cite{ke2017lightgbm}. The main difference between XGBoost and LightGBM is how they grow the decision tree (see Figure \ref{fig: modelscomparison} for a high-level comparison). In XGBoost decision trees are grown horizontally (level-wise) while with LightGBM decision trees are grown vertically (leaf-wise). Importantly, this makes LightGBM an effective algorithm to handle datasets with high dimensionality.

\subsubsection{Voting Ensemble} \label{voting ensemble}

\begin{figure}[H]
\centering
\includegraphics[width=3.6in]{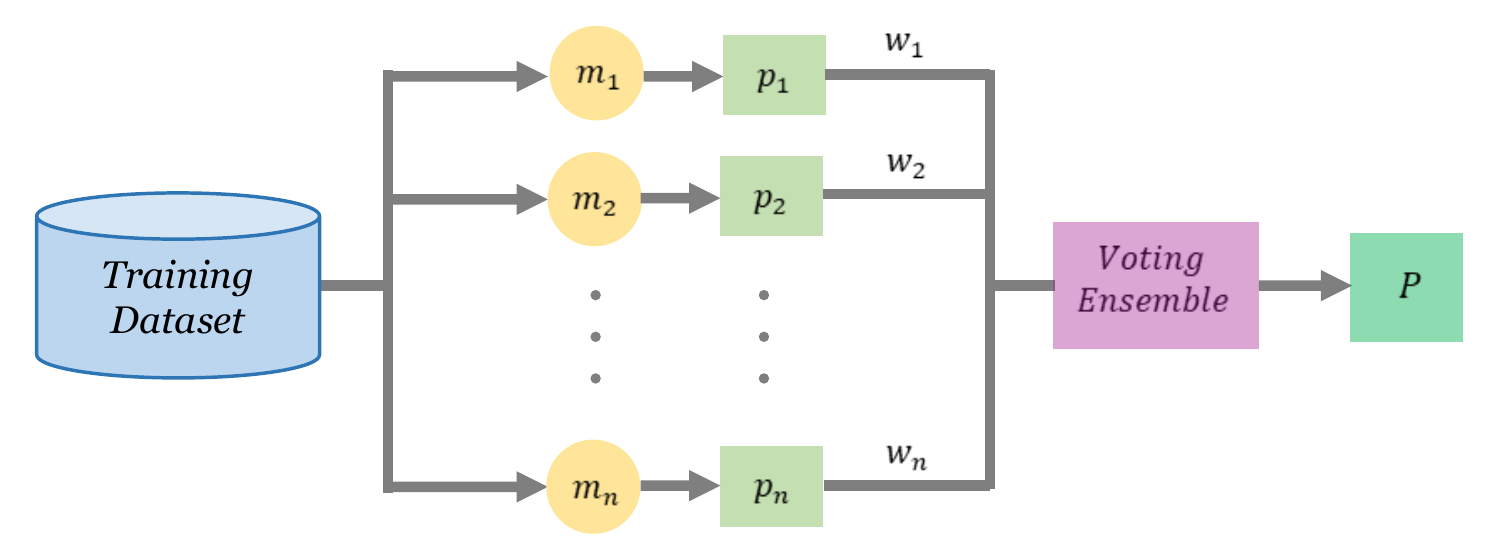}
\caption{Voting Ensemble combines predictions \(p_i\) of multiple classifiers \(m_i\) using weighted average \(w_i\) to compute a final prediction \(P\).}
\label{fig: voting ensemble}
\end{figure}

A voting ensemble was used to integrate the predictions of multiple iterations of both XGBoost and LightGBM classifiers with different parameterizations. Specifically, a soft voting weighted average voting ensemble was used to combine the predictions for each model (Figure \ref{fig: voting ensemble}). A soft voting ensemble is a meta-classifier model which takes the weighted average probability predicted by each classifier:

\begin{equation} \label{eqn8}
    \hat{y_i} = \textrm{argmax}_i \sum_{j=1}^{m} w_j p_{ij} 
\end{equation}

\subsection{Workflow and Pipelines}
\unskip

Pipeline is defined as an executable workflow of data that is encapsulated in a series of steps. In this work, the proposed workflow consists of two main pipelines: (1) ML pipeline and (2) Prediction pipeline. All pipeline codes were custom-written in Python 3.7 on Microsoft Azure Machine Learning Service \cite{barga2015introducing} cloud platform. XGBoost v1.1 and LightGBM v2.3.1 libraries were integrated into Python to create ML classification models. The voting ensemble was created using the Microsoft Automated Machine Learning tool \cite{feurer2015efficient}.

\subsubsection{Machine Learning Pipeline}
\unskip

\begin{figure}[H]
\centering
\includegraphics[width=4.2in]{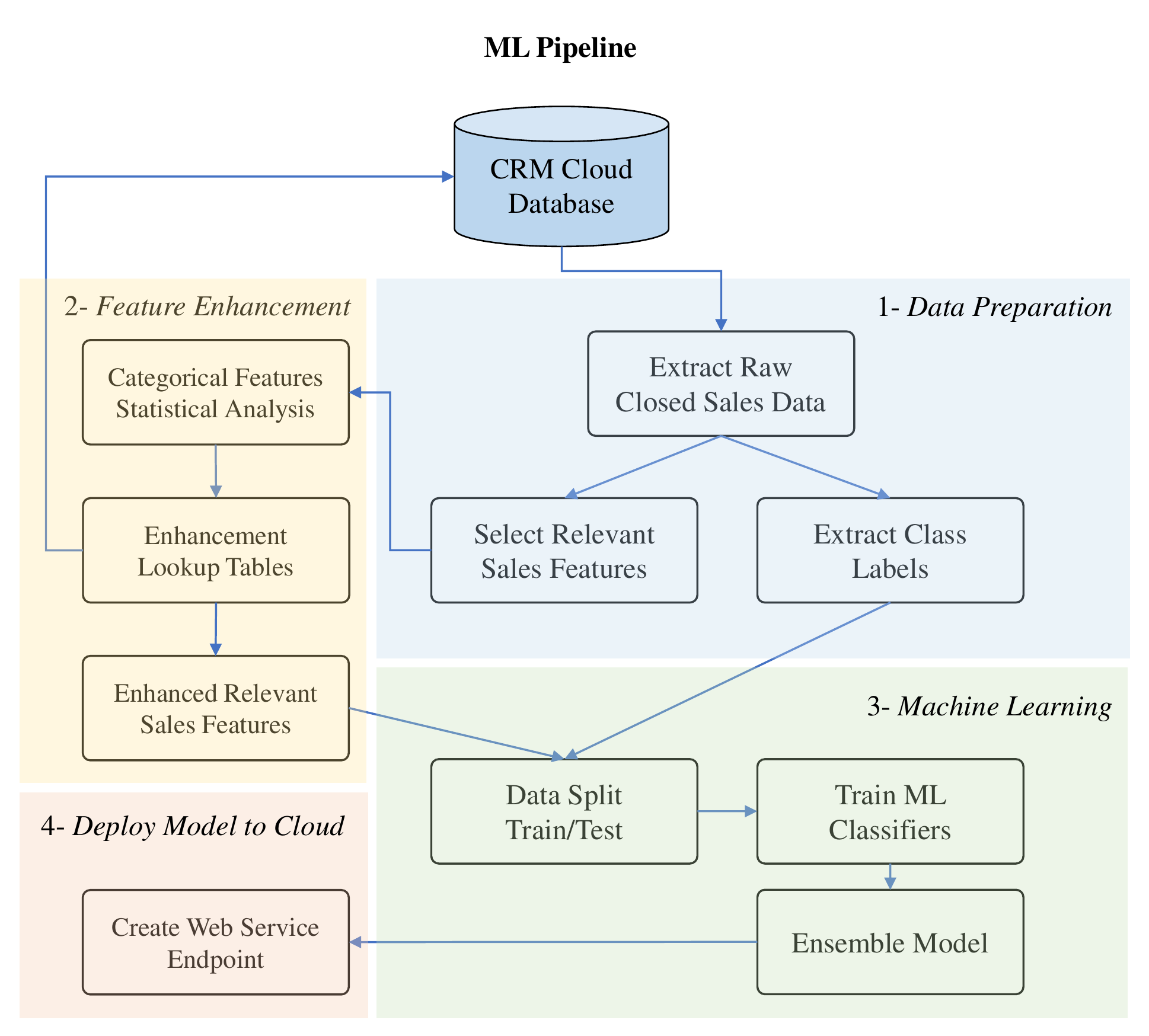}
\caption{ML Pipeline: In four major steps the pipeline extracts and enhances sales data from a cloud database, trains an ensemble of ML classification models on the data, and eventually creates a cloud endpoint for the model.}
\label{fig: train pipeline}
\end{figure}

The main objective of the ML pipeline is to train predictive models on the closed sales opportunities data. As illustrated in Figure \ref{fig: train pipeline}, there are four main steps in this pipeline:

(1) \textit{Data Preparation}: Raw data of all closed sales opportunities are extracted from the CRM cloud database. Relevant features are selected for each sales record (see Table \ref{table: features}) and paired with their sales outcome (won/lost) as a class label. Note that the user-entered probabilities are dropped to avoid biasing the model's predictions.

(2) \textit{Feature Enhancement}: As described in section \ref{Feature Enhancement}, statistical analysis is performed on all categorical features to generate feature enhancement lookup tables for each of these categorical features (see Figure \ref{fig: lookup table}). All lookup tables are stored back in the CRM cloud database. These tables are then appropriately merged back to the original selected features in the raw data.

(3) \textit{Machine Learning}: A total number of 35 iterations of XGBoost and LightGBM classifiers with various parameterizations are trained on the data (section \ref{XGBoost and LightGBM}). Eventually, all trained models are combined to generate a soft-voting ensemble classifier (section \ref{voting ensemble}).

(4) \textit{Deploy Model to Cloud}: In the last step of the ML pipeline, the ensemble model is deployed as a web service using Azure ML. Azure ML platform supports creating a model's endpoint on Azure Kubernetes Service (AKS) cluster \cite{barnes2015azure}. AKS enables request-response service with low latency and high scalability which makes it suitable for production-level deployments.

\subsubsection{Prediction Pipeline}\label{prediction pipeline}
\unskip

\begin{figure}[H]
\centering
\includegraphics[width=4.8in]{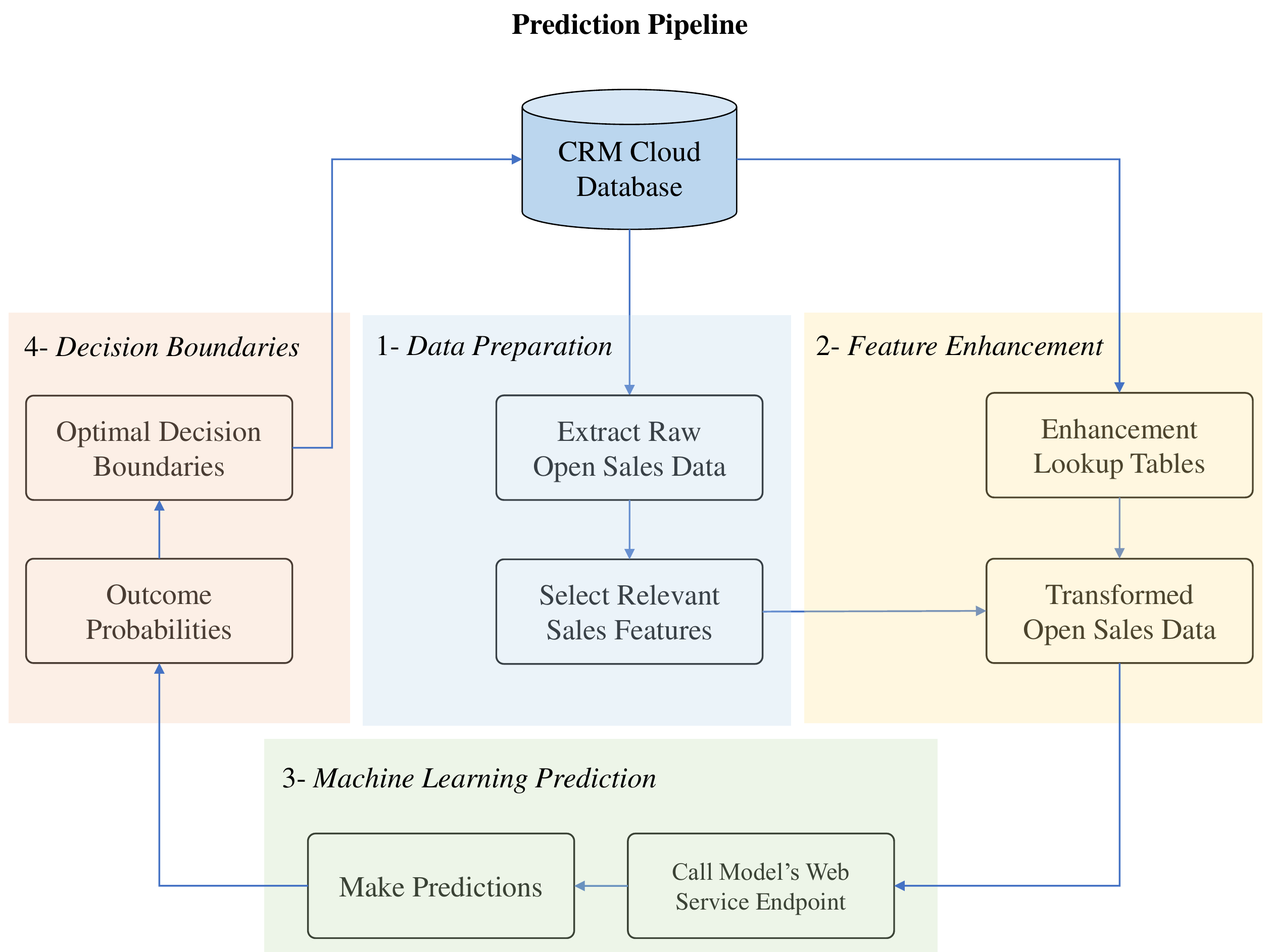}
\caption{Prediction Pipeline: new sales opportunities data are transformed and enhanced, probability of winning is inferred using the trained ML model, and finally decision boundaries are optimized based on historical sales data.}
\label{fig: test pipeline}
\end{figure}

The prediction pipeline, as illustrated in Figure \ref{fig: test pipeline}, was designed to utilize the trained ML model and make predictions on the likelihood of winning new sales opportunities in four main steps:

(1) \textit{Data Preparation}: All open sales records are extracted from the CRM cloud database. Relevant features are selected similar to the feature selection step in the ML pipeline. Note that open sales opportunities are still active in the sales process and, hence, there is no final sales status (won/lost) assigned to them yet.

(2) \textit{Feature Enhancement}: To make predictions on unseen data using the trained ML model, new input data needs to have a format similar to the data used to train the model. Therefore, all the previously stored lookup dictionaries are imported from the CRM cloud database and appropriately merged to the relevant features.

(3) \textit{Machine Learning Prediction}: The ensemble model created in the ML pipeline is called using its endpoint. The model makes predictions on the unseen sales data and assigns a probability of winning to each new opportunity.

(4) \textit{Decision Boundaries}: All historical data on closed sales opportunities along with their ML predicted probabilities are grouped by the business segments (Healthcare, Energy, and Finance). Next, within each business segment, closed sales records are split into four quartiles based on their value. Then, the optimal decision boundary is calculated for each business segment's value quartile as described in section \ref{binary classification}. A total number of 12 decision boundaries are calculated (3 business segments \(\times\) 4 quartiles). Eventually, all predicted probabilities and decision boundaries are stored back to the cloud database.

%%%%%%%%%%%%%%%%%%%%%%%%%%%%%%%%%%%%%%%%%%
\section{Results}

This section gives an overview of the proposed workflow's performance. The workflow was implemented in the CRM system of a global B2B consulting firm. The two pipelines were scheduled for automated runs on a recurring basis on the Azure ML platform. The ML pipeline was scheduled for a weekly rerun to retrain ML models on updated sales data and generate updated feature enhancement lookup tables. The prediction pipeline was scheduled for a daily rerun to calculate and store predictions for new sales opportunities.

%%%%%%%%%%%%%%%%%%%%%%%%%%%%%%%%%%%%%%%%%%
\subsection{Training the ML Model}

\begin{figure}[H]
\centering
\includegraphics[width=4.9in]{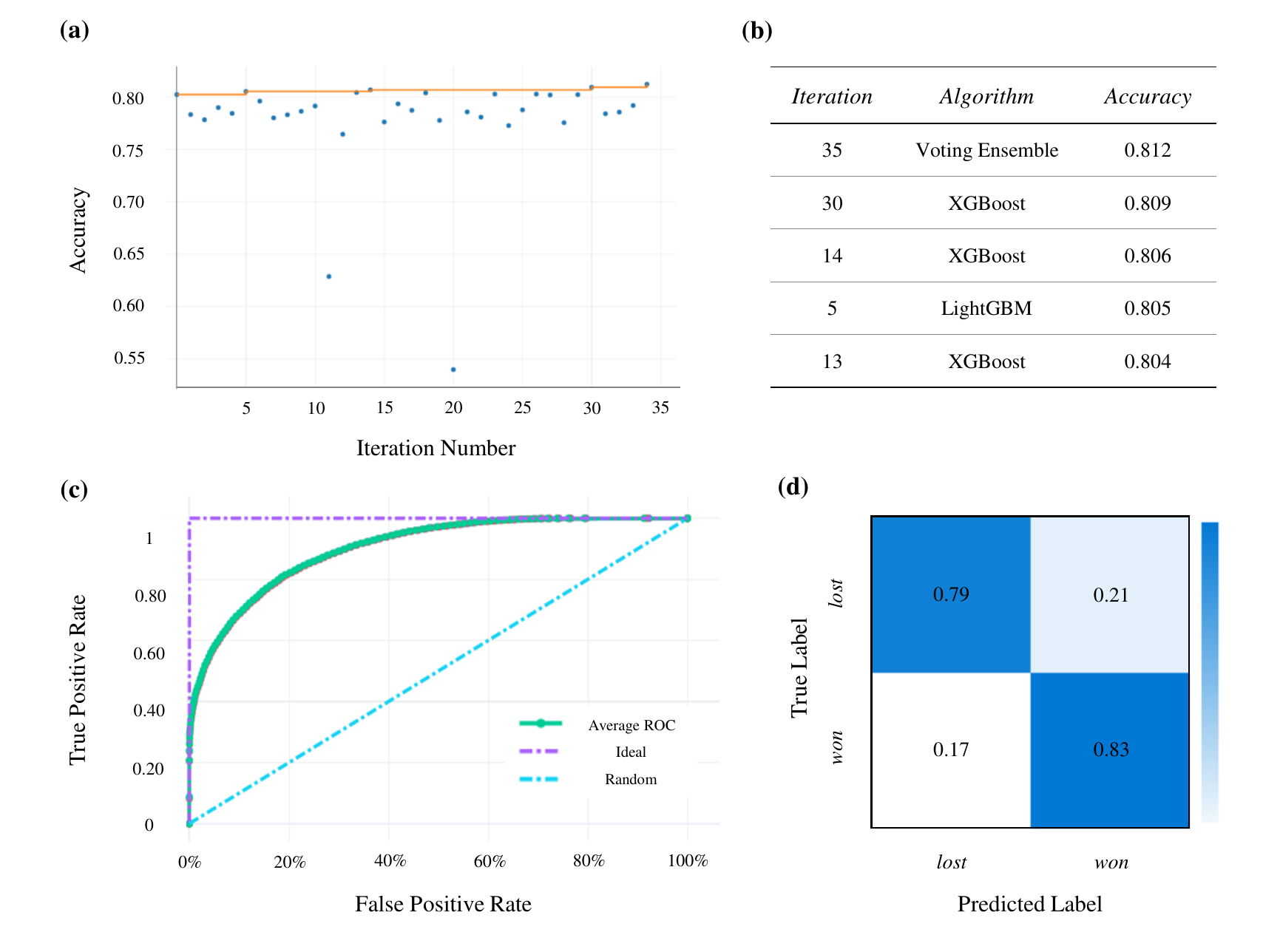}
\caption{ML Training Results: (\textbf{a}) Training accuracy for all model iteration. (\textbf{b}) Accuracy of the top five iterated models sorted by the training accuracy. (\textbf{c}) ROC curve of the voting ensemble classifier. (\textbf{d}) Confusion matrix showing the four scenarios of classification for the voting ensemble model. }
\label{fig: train performance}
\end{figure}

A total number of 34 iterations of XGBoost and LightGBM were individually trained on the data and then combined in a voting ensemble classifier (see section \ref{voting ensemble} for more details). The training accuracy was calculated using 10-fold cross-validation. The accuracy for each of the 35 iterations (with the last iteration being the voting ensemble classifier) is demonstrates in Figure \ref{fig: train performance}-A. Training accuracy for the top five model iterations are listed in Figure \ref{fig: train performance}-B. As expected, the voting ensemble had a slightly higher training accuracy compared to each individual classifier.

The voting ensemble classifier had a training accuracy of \(81\%\) (other performance metrics are listed in Table \ref{table: train performance}). On the train set, approximately \(83\%\) of the won sales and \(79\%\) of the lost sales were classified correctly (Figure \ref{fig: train performance}-D). For more insight into the training performance ROC curve (Figure \ref{fig: train performance}-C) is also illustrated. The area under the ROC curve (AUC) was equal to \(0.87\). In other words, this implies that a randomly drawn sample out of the train set has a \(87\%\) chance of being correctly classified by the trained model.

\begin{table}
\centering
\caption{Voting Ensemble Training Performance.}
\begin{tabular}{ll} 
\toprule
\textbf{Metric} & \textbf{Value}  \\ 
\midrule
Precision       & 0.81           \\
Recall          & 0.83           \\
Accuracy        & 0.82           \\
AUC             & 0.87            \\
\bottomrule
\end{tabular}
\label{table: train performance}
\end{table}

\subsection{Setting the Decision Boundaries}

\begin{figure}[H]
\centering
\includegraphics[width=3.5in]{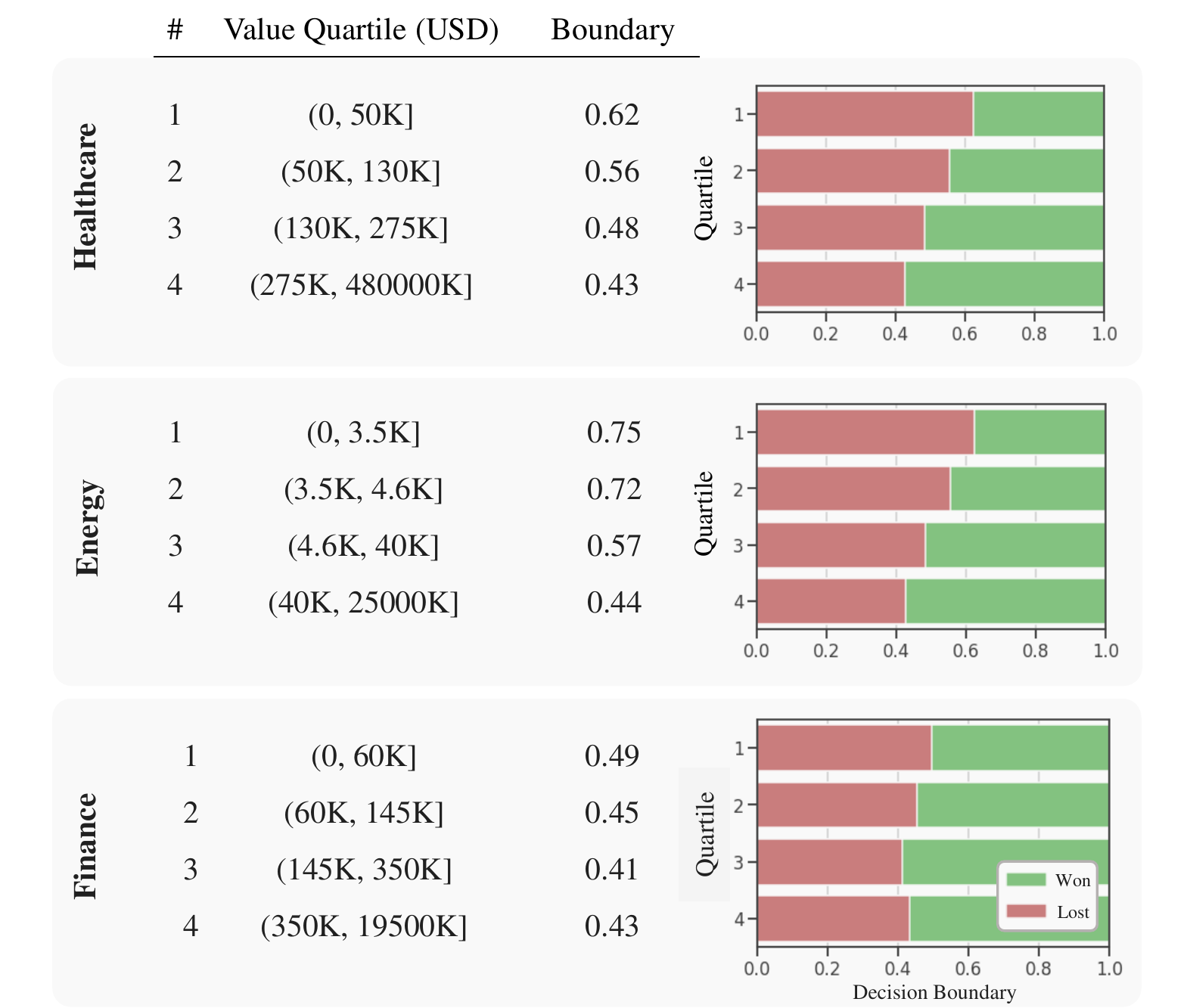}
\caption{Decision Boundaries: .}
\label{fig: decision boundaries segment quartile}
\end{figure}

As explained in section \ref{prediction pipeline}, statistical analysis of historical sales data is performed in each business segment (Healthcare, Energy, and Finance) to determine the decision boundaries. Specifically, the decision boundary was optimized for each of the four sales value quartiles of each business segment. The decision boundaries, demonstrated in Figure \ref{fig: decision boundaries segment quartile}, ranged from \(0.41\) (Finance business segment - 3\textsuperscript{rd} value quartile) to \(0.75\) (Energy business segment - 1\textsuperscript{st} value quartile).

Interestingly, the decision boundaries were lower for sales opportunities with a higher monetary value which implies a more optimistic decision making for more profitable opportunities. This sensible trend observed in the optimal decision boundaries provides more evidence to substantiate the idea of tailoring the boundaries uniquely to each business segment and value quartile due to their inherent decision-making differences.

\subsection{Model's Performance}

The voting ensemble was used to make predictions on the unseen test set. In particular, after inferring the probability of winning for each sales opportunity, they were classified in accordance with a decision boundary corresponding to their business segment and value quartile. If the inferred probability of winning exceeded the decision boundary, a sales opportunity was classified to be won otherwise it was classified to be a lost opportunity. To make a concrete comparison between user-entered and ML predictions, statistical and monetary performance metrics were calculated for both approaches.

\begin{figure}[H]
\centering
\includegraphics[width=3.9in]{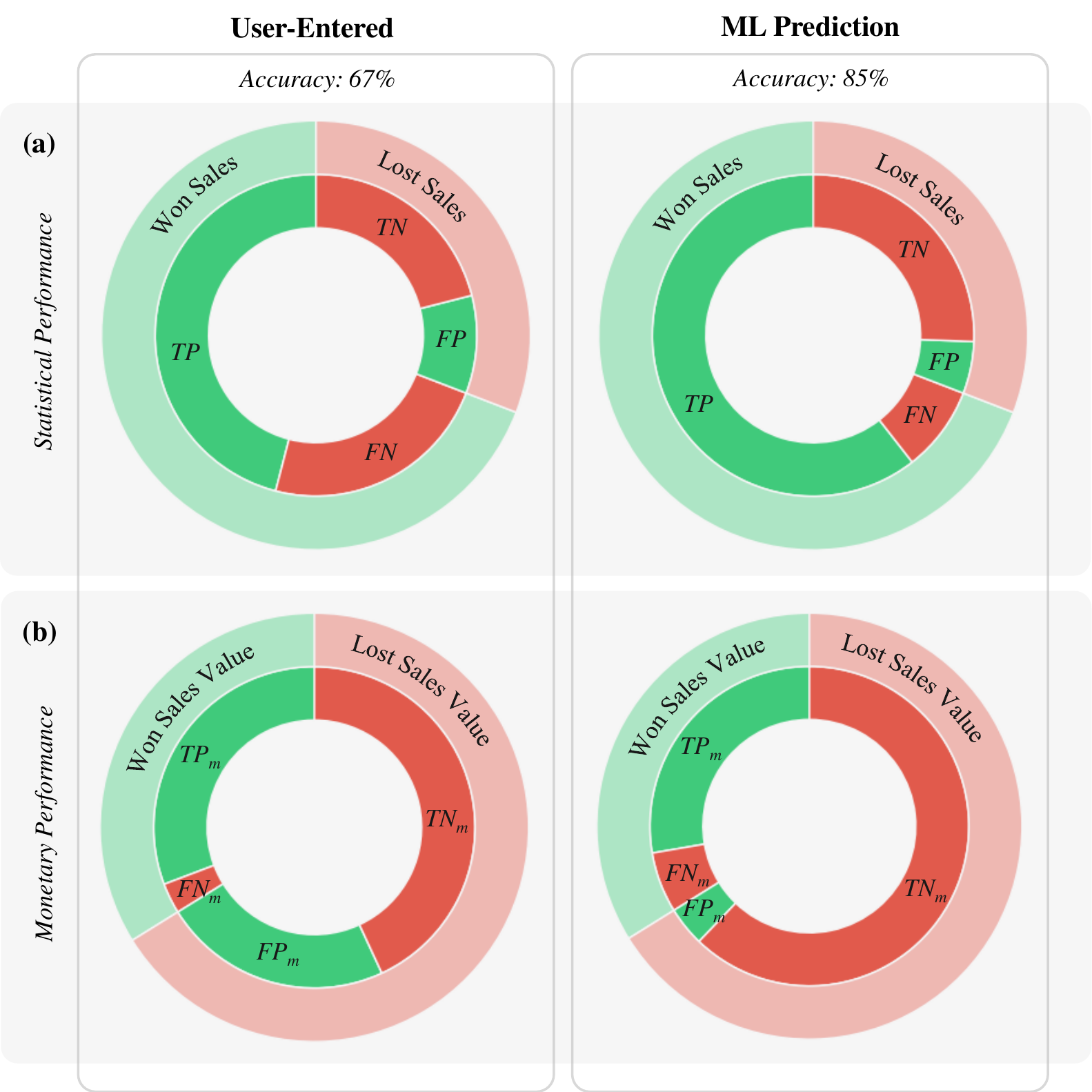}
\caption{Test Set Results: (\textbf{a}) Statistical Performance: Actual outcome of all won (light green) and lost (light red) sales opportunities along with the corresponding predictions (solid green and red). (\textbf{b}) Monetary Performance: Cumulative contract value of won (light green) and lost (light red) sales opportunities along with the cumulative value of opportunities in each of the four classification scenarios (solid green and red). In both panels miss-matching colors indicates false classification.}
\label{fig: test set results}
\end{figure}

All four classification scenarios in the test set for both user-entered and ML prediction are depicted in Figure \ref{fig: test set results}-A. Qualitatively, the ML workflow made fewer false classifications (i.e, compare the true positive \(TP\) slice proportions in Figure \ref{fig: test set results}-A). More specifically, the ML workflow accurately classified \(87\%\) of the unseen sales data while the user-entered predictions only had an accuracy of \(67\%\). In fact, all statistical performance metrics (precision, recall, and F1 score) were in favor of the ML predictions (see Table \ref{table: test performance}).

The performance of the user-entered and ML predictions was also compared with reference to the monetary metrics (see section \ref{binary classification} for more details). As shown in Figure \ref{fig: test set results}-B, sales opportunities falsely predicted to be won by the ML workflow had considerably lower cumulative monetary value (compare the true positive \(FP_m\) slice proportions). This implies a lower monetary loss due to prediction error when using the ML predictions. Quantitatively, the monetary accuracy of the ML model was notably higher than the user-entered (\(90\%\) versus \(74\%\)). Other monetary performance metrics are listed in Table \ref{table: test performance}.

\begin{table}
\centering
\caption{Test Set Performance Metrics}
\begin{tabular}{lcc} 
\toprule
\textit{Statistical Performance} &      &               \\ 
\midrule
\textbf{Metric}                 & \textbf{User-Entered}   & \textbf{ML}   \\ 
\midrule
Precision               & 0.82 & 0.92           \\
Recall                  & 0.66 & 0.87           \\
F1-Score                & 0.73 & 0.89           \\
Accuracy                & 0.67 & 0.85           \\ 
\toprule\toprule
\textit{Monetary Performance}    &      &               \\ 
\midrule
\textbf{Metric}                  & \textbf{User-Entered}  & \textbf{ML}   \\ 
\midrule
Precision\_m            & 0.57 & 0.87           \\
Recall\_m               & 0.91 & 0.82           \\
Accuracy\_m             & 0.74 & 0.90           \\
\bottomrule
\end{tabular}
\label{table: test performance}
\end{table}

\subsection{Analysis of the Workflow Implementation}

\begin{figure}[H]
\centering
\includegraphics[width=3.9in]{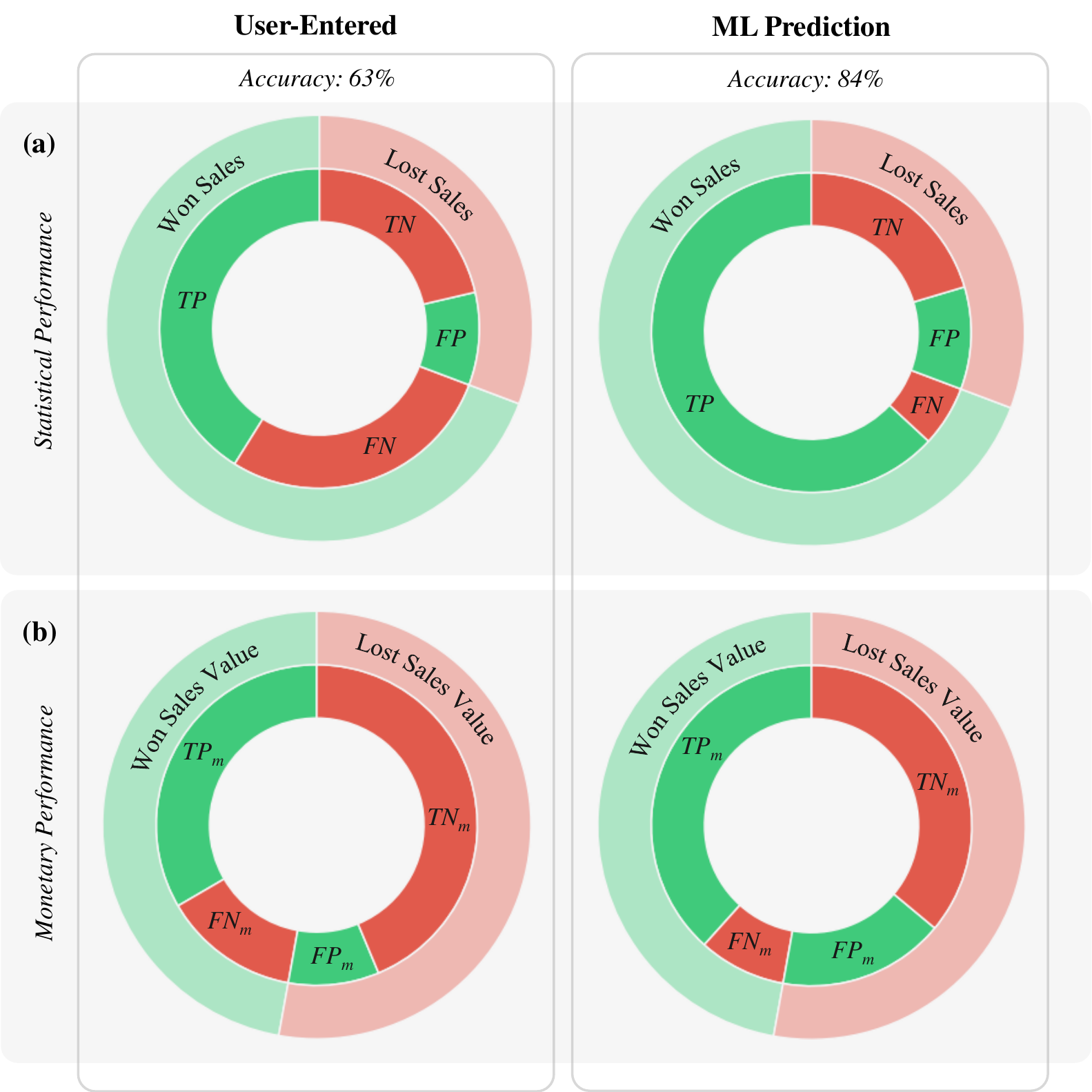}
\caption{Validation Set Results: (\textbf{a}) Statistical and (\textbf{b}) Monetary performances of user-entered and ML predictions. Refer to Figure \ref{fig: test set results} caption for further explanations.}
\label{fig: valid set results}
\end{figure}

Similar to the previous section, a performance comparison between the user-entered and ML predictions was performed on a validation set. The validation set was collected while the workflow was implemented in the sales pipeline of a B2B consulting firm over a period of three months (see section \ref{data and feature} for further details). A qualitative comparison in terms of statistical and monetary performance is presented in Figure \ref{fig: valid set results}. In the validation set, the ML workflow retained a substantially higher prediction accuracy (\(83\%\) versus \(63\%\)). Also, there was an evident gap between the number of won sales misclassified by each approach (compare the true positive \(TP\) slices in Figure \ref{fig: valid set results}-A).

The monetary accuracy of the ML predictions was marginally lower than the user-entered predictions (\(75\%\) versus \(77\%\)). However, the cumulative value of the won sales opportunities correctly classified by the ML workflow was still considerably higher than the user-entered predictions (compare the true positive \(TP_m\) slices in Figure \ref{fig: valid set results}-B). All performance metrics are listed in Table \ref{table: valid performance}.

\begin{table}
\centering
\caption{Validation Set Performance Metrics}
\begin{tabular}{lcc} 
\toprule
\textit{Statistical Performance} &      &               \\ 
\midrule
\textbf{Metric}                  & \textbf{User-Entered} &\textbf{ML}     \\ 
\midrule
Precision                & 0.82  & 0.85        \\
Recall                   & 0.59  & 0.92        \\
F1-Score                 & 0.70  & 0.87         \\
Accuracy                 & 0.63  & 0.83        \\ 
\toprule\toprule
\textit{Monetary Performance}    &      &               \\ 
\midrule
\textbf{Metric}                  & \textbf{User-Entered}  & \textbf{ML}   \\ 
\midrule
Precision\_m            & 0.79 & 0.71           \\
Recall\_m               & 0.72 & 0.82           \\
Accuracy\_m             & 0.77 & 0.75           \\
\bottomrule
\end{tabular}
\label{table: valid performance}
\end{table}

%%%%%%%%%%%%%%%%%%%%%%%%%%%%%%%%%%%%%%%%%%
\section{Conclusions}

In this paper, we proposed a novel ML workflow implemented on a cloud platform for predicting the likelihood of winning sales opportunities. With this approach, sales data were extracted from the CRM cloud database and then improved by an extensive feature enhancement approach. The data was then used to train an ensemble of probabilistic classification models in parallel. The resulting meta classification model was then used to infer the likelihood of winning new sales opportunities. Lastly, to maximize the interpretability of the predictions, optimal decision boundaries were calculated by performing statistical analysis on the historical sales data.

To inspect the effectiveness of the ML approach, it was deployed to a multi-business B2B consulting firm for over three months. The performance of the ML workflow was compared with the user-entered predictions made by salespersons. Standard statistical performance metrics confirmed that by far the ML workflow provided superior predictions. From a monetary standpoint, the value created from decision-making was also higher when incorporating the ML workflow.

The proposed ML workflow is a cloud-based solution that can readily be integrated into the existing cloud-based CRM system of enterprises. On top of that, this workflow is highly sustainable and scalable since it relies on cloud computing power instead of on-premise computing resources.

The results obtained in this work suggest a data-driven ML solution for predicting the outcome of sales opportunities is a more concrete and accurate approach compared to salespersons' subjective predictions. However, it is worth mentioning that ML solutions should not be overwhelmingly used to rule out sensible or justifiable sentiments of salespersons in evaluating a sales opportunity. A data-driven approach, such as the workflow presented in this work, can provide a reliable reference point for further human assessments of the feasibility of a sales opportunity.

%%%%%%%%%%%%%%%%%%%%%%%%%%%%%%%%%%%%%%%%%%
\vspace{6pt} 

%%%%%%%%%%%%%%%%%%%%%%%%%%%%%%%%%%%%%%%%%%
%% optional
%\supplementary{The following are available online at \linksupplementary{s1}, Figure S1: title, Table S1: title, Video S1: title.}

%%%%%%%%%%%%%%%%%%%%%%%%%%%%%%%%%%%%%%%%%%
%% optional
%\abbreviations{The following abbreviations are used in this manuscript:\\
%
%\noindent 
%\begin{tabular}{@{}ll}
%MDPI & Multidisciplinary Digital Publishing Institute\\
%DOAJ & Directory of open access journals\\
%TLA & Three letter acronym\\
%LD & linear dichroism
%\end{tabular}}

%%%%%%%%%%%%%%%%%%%%%%%%%%%%%%%%%%%%%%%%%%
\reftitle{References}


\externalbibliography{yes}
\bibliography{Ref}

%=====================================
% References, variant B: internal bibliography
%=====================================
%\begin{thebibliography}{999}
% Reference 1

%\bibitem[Author1(year)]{ref-journal}
%Author1, T. The title of the cited article. {\em Journal Abbreviation} {\bf 2008}, {\em 10}, 142--149.
% Reference 2
%\bibitem[Author2(year)]{ref-book}
%Author2, L. The title of the cited contribution. In {\em The Book Title}; Editor1, F., Editor2, A., Eds.; Publishing House: City, Country, 2007; pp. 32--58.

%\end{thebibliography}

% The following MDPI journals use author-date citation: Arts, Econometrics, Economies, Genealogy, Humanities, IJFS, JRFM, Laws, Religions, Risks, Social Sciences. For those journals, please follow the formatting guidelines on http://www.mdpi.com/authors/references
% To cite two works by the same author: \citeauthor{ref-journal-1a} (\citeyear{ref-journal-1a}, \citeyear{ref-journal-1b}). This produces: Whittaker (1967, 1975)
% To cite two works by the same author with specific pages: \citeauthor{ref-journal-3a} (\citeyear{ref-journal-3a}, p. 328; \citeyear{ref-journal-3b}, p.475). This produces: Wong (1999, p. 328; 2000, p. 475)

%%%%%%%%%%%%%%%%%%%%%%%%%%%%%%%%%%%%%%%%%%

%% for journal Sci
%\reviewreports{\\
%Reviewer 1 comments and authors’ response\\
%Reviewer 2 comments and authors’ response\\
%Reviewer 3 comments and authors’ response
%}

%%%%%%%%%%%%%%%%%%%%%%%%%%%%%%%%%%%%%%%%%%

\end{document}